# SMART GRID: A SURVEY OF ARCHITECTURAL ELEMENTS, MACHINE LEARNING AND DEEP LEARNING APPLICATIONS AND FUTURE DIRECTIONS

Navod Neranjan Thilakarathne[1], Mohan Krishna Kagita[2] and Dr.Thippa Reddy Gadekallu[3]

[1] Department of ICT, Faculty of Technology, University Of Colombo, Srilanka
[2] School of Computing and Mathematics, Charles Sturt University, Melbourne, Australia
[2] School of Information Technology and Engineering, Vellore Institute of Technology, INDIA
`navod.neranjan@ict.cmb.ac.lk`, `mohankrishna4k@gmail.com`,
`thippareddy.g@vit.ac.in`

**Abstract.** The Smart grid (SG), generally known as the next-generation power grid emerged as a replacement for ill-suited power systems in the 21st century. It is integrated with advanced communication and computing capabilities, thus it is expected to enhance the reliability and the efficiency of energy distribution with minimum effects. With the massive infrastructure it holds and the underlying communication network in the system, it introduced a large volume of data that demands various techniques for proper analysis and decision making. Big data analytics, machine learning (ML), and deep learning (DL) plays a key role when it comes to the analysis of this massive amount of data and generation of valuable insights. This paper explores and surveys the Smart grid architectural elements, machine learning, and deep learning-based applications and approaches in the context of the Smart grid. In addition in terms of machine learning-based data analytics, this paper highlights the limitations of the current research and highlights future directions as well.

**Keywords:** Smart Grid, Internet of Things, Machine Learning, Deep Learning, Cyber Physical System

## 1 Introduction

Over the years there has nothing changed in the basic structure of the electrical power grid and it is observed that it has become obsolete and ill-suited [5,6], that it cannot cater to the demand and supply in the 21st century [1]. Even though we are in the 21st century, electrical infrastructure has remain unchanged over the years. But the demand for electricity has being risen with the increased consumption and the population. According to the U.S department of Energy, they have stated that consumption and the demand for electricity in the U.S have increased 2.5% annually over the last twenty years [1].This traditional energy distribution approach is ill-suited for catering to the increasing demand and it has many deficiencies such as poor and slow response time, lack of on demand analysis and lack of situational awareness which ultimately caused to the blackouts happening over past years. In order to overcome the challenges of existing obsolete electrical power grid, the new smart grid concept has emerged. This is regarded as the next generation electrical power grid that uses two way flows of electricity and information to create an automated and distributed energy delivery network [2].This SG is also called as intelligent grid and intelligrid. The conventional power grid supplied power to a large number of customers from the central power plant or from central generators, while SG uses bidirectional information and electricity flows to create an integrated and distributed advanced energy supply network. The modern SG relies heavily on the Internet of Things (IoT) [39], which acts as its primary enabler. It also relies heavily on Cyber Physical Systems (CPS) that can track, exchange, and handle information and activities and serve as an integral part of the SG [39]. Cyber-Physical Systems (CPS) are a mixture of computational and physical characteristics and can be widely found in different domains, including the modern Smart grid.

Basically, SG is a vast electricity network that uses real-time and intelligent monitoring, communication, control, and self-healing technologies enabling versatile options for prosumers ensuring electricity supply reliability and protection.SGs are a complex Cyber Physical System by their very nature. It features a centralized approach where few central stations distribute energy to various consumers [39]. Over the years, energy systems have developed from traditional power systems to SGs and CPS's in energy [42].

It is no doubt that electricity is vital for the existence of human life, hence this SG requires high reliability where the faith of the economies are heavily dependent on the existence of this SG's [40]. The functionality of this modern SG can be broken down into four parts.
1. Generation: In this phase electricity produces in many ways. (e.g. burn of fossil fuels, wind, solar, nuclear reaction)
2. Transmission: Transmit electricity through a very high voltage electronic infrastructure
3. Distribution: Spread out electricity for further distribution
4. Consumption: Various industries, residents are using electricity for various purposes

When it comes to the traditional grid outages are often recognized only after consumers report [40]. It is difficult to match generation to demand because utilities do not have clear cutting methods to forecast demand and request demand reduction. As a result, they need to over generate electricity, which is costly and leads to greenhouse gas emissions, for peak demand. Because of the variable generation, such as wind and solar power, is hard to integrate into the grid.

SG offers several advantages. Using intelligent communications, load shedding that can be implemented to flatten peak demand, reducing the need for additional generation plants to be brought online. Using ML and DL to conduct predictive analysis, including when less power is generated by wind and solar resources, the utilities may keep power adequately balanced. As new storage technologies emerge on the utility scale, intelligent demand prediction would also benefit from the inclusion of these devices. Finally, the ability to receive and respond to price signals for customers would help them control their energy costs, thus helping utilities prevent the construction of additional generation plants. The smart grid makes a dramatic reduction in costs for both power generation and consumption from all these approaches [40].

## 1.1 Internet of Things in the Smart Grid

The ongoing rapid advancement of the Internet of Things [33,35,36,37,38,43], provide solutions to overcome the emerging challenges of SG. The SG use IoT enabled technologies for creating various intelligent services that are essential for the functionality of SG [41]. The development aspect of most of the SG has enhanced by the applications of IoT enabled technologies.

The IoT architecture of SG comprises of three main layers. That is

1. Perception layer
2. Network layer
3. Application layer

These IoT solutions are used for various purposes, such as monitoring of power transmission lines, smart homes, management of electric vehicles, and smart patrols. In addition, IoT technologies play an important role in SG, such as network design, network security management, operation and maintenance, data collection, security monitoring, measurement, and user interaction. Potential IoT applications can be extended to all facets of the power system, including generation, transmission, transformation, distribution and utilization of electricity [41].

## 1.2 The Basic Architecture of SmartGrid

The modern Smart grid is consist of automated control, modern energy management techniques, novel communication infrastructure, power converters, sensing and measuring technologies and the underlying infrastructure capable of facilitating for reliable and efficient delivery of energy. Importantly this new smart grid is much different and complex than the traditional one. The traditional grid lacks communication capabilities whereas this modern smart grid infrastructure comprises of advanced controls and distributed delivery system as depicted on Fig. 1.

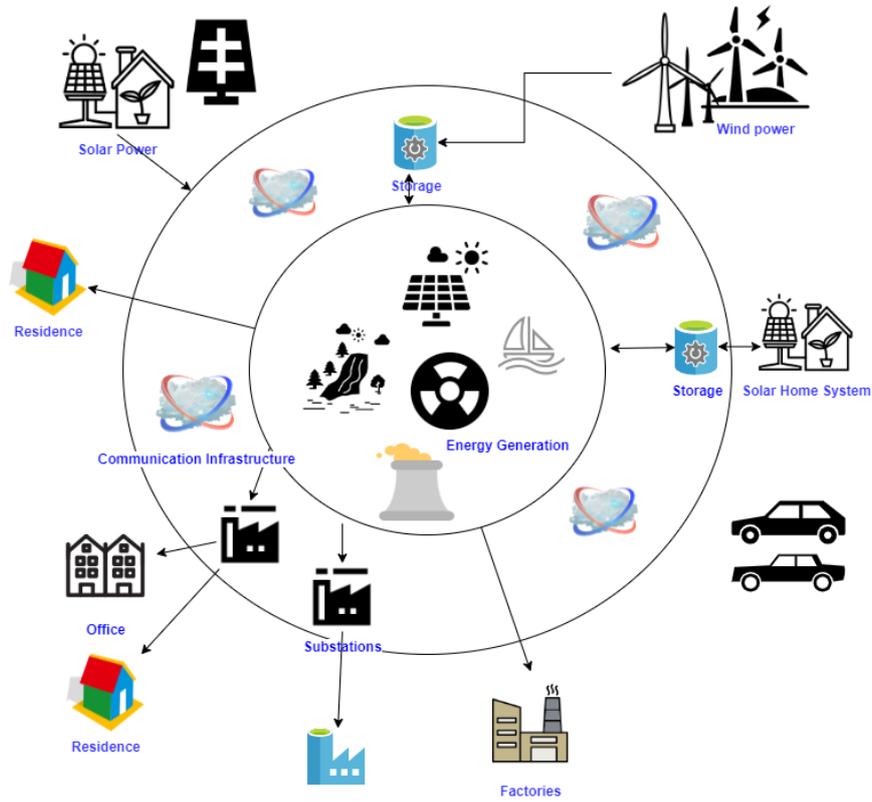

**Fig. 1.** A Basic Architecture of Smart Grid.

As depicted on the Fig. 1 various components of the grid system are linked together with sensing nodes and communication paths to provide interoperability between them. In the SG real time information and data analytics becomes the key factor for ensuring a consistent and reliable delivery of energy, from energy generating units to the customers. In case of equipment failures, natural disasters and power disturbances and outages SG provide protection by real-time condition monitoring, real-time diagnostic and outage monitoring. Especially for the existence of SG, automated control and for billing purposes secure information transportation and storage is highly required. As many number of Internet connected devices are present and connected to the communication network, forming a large IoT network efficient security mechanism should be devised in order to avoid malicious cyber-attacks.

In the second section of this paper, we describe various aspects of machine learning and deep learning applications for SG. Next section we discuss the future directions and next we describe about the limitations of the current research. Finally, we conclude our paper with the conclusion and the summary of the study.

## 2     Machine Learning and Deep Learning Applications for SmartGrid

ML and DL appeared as an innovative tool for analysis of the Big Data generated by the underlying IoT devices, critical infrastructure, and the massive communication network in order to properly analyze and make timely decisions to run the SG. Big Data refers to the massive amount of data that needs to be captured, curated, managed, and analyzed by more advanced methods [8]. Machine learning is a term that refers to the continuous learning and predictions from the data available. It consists of different algorithms that analyze the data to produce decisions or predictions regarding the current context through a set of instructions. Deep Learning is a subfield of machine learning that deals with algorithms inspired by the structure and function of the brain called artificial neural networks [28,29,30]. ML and DL functionalities in the context of SG includes predicting about,

1. Energy generation
2. Price
3. Energy consumption
4. Fault detection
5. Sizing
6. Network anomaly detection
7. Security breach detection
8. Fraud detection
9. Optimum schedule
10. Stability of the SG

Fig. 2 depicts the various applications of Machine Learning.

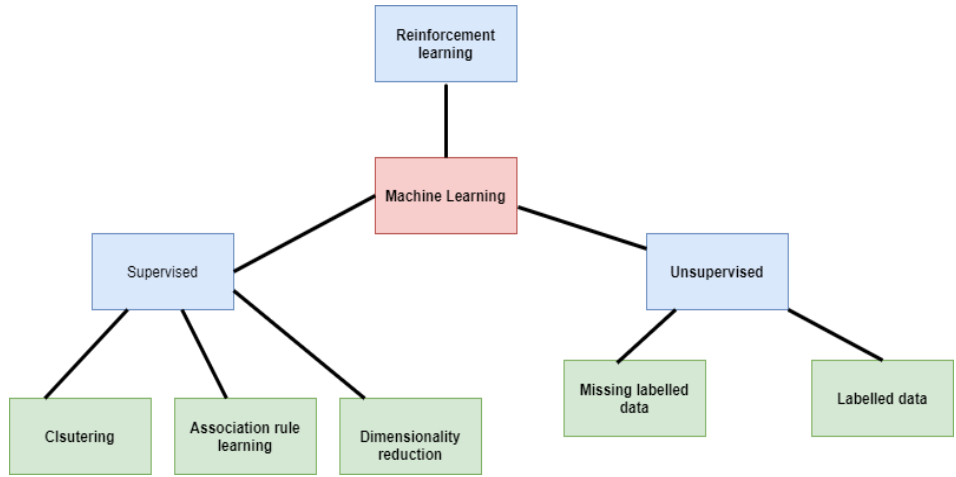

**Fig. 2.** Applications of ML

In the next subsections, we discuss key application areas of ML and DL in the context of SG. That is energy forecasting and protection and the security of SG. Next subsection we discuss the rest of the ML and DL application areas for SG.

### 2.1  ML and DL Applications in Energy Forecasting

Solar energy is considered as one of the major renewable energy source which is actually threatened by many impediments such as seasonal changes, weather condition, topographic elevation and etc. As a result, clear forecasting is required to obtain accurate details of energy generation. Further wind power is also considered as a rapidly growing energy source. Same like solar power, wind power poses many obstacles especially in power generation and transmission, such as varying wind speed which leads to fluctuation of the output in the wind power plant. Such things will lead to instability of the SG and require accurate forecasting, where this ML and DL forecasting models come in to play. Following Table 1 summarizes the ML and DL applications that are introduced and devised for SG energy forecasting.

**Table 1.** ML and DL Applications for Energy Forecasting

| Reference | Year | ML/DL Technique | Application |
| --- | --- | --- | --- |
| [44] | 2006 | Adaptive neuro-fuzzy inference | Prediction of short term wind for power generation |
| [45] | 2006 | Recurrent neural network | Deals with the problem of long-term wind speed and power forecasting based on meteorological information |
| [46] | 2012 | Fuzzy modeling | A short term wind farm power output prediction model |
| [47] | 2008 | Artificial neural network | Analysis of wind power generation and prediction |
| [48] | 2013 | k-nearest neighbor (k-NN) classification | Deals with very short term wind speed prediction |
| [49] | 2013 | Least median square (LMS), multilayer perceptron (MLP) and support vector machine (SVM) | Deals with the improving the accuracy for solar power prediction |
| [50] | 2014 | Extreme Learning Machine | Prediction of the daily global solar radiation |
| [51] | 2016 | Artificial neural networks (ANN) and support vector regression (SVR) | Predicting energy productions from a solar photovoltaic (PV) system |
| [52] | 2017 | Regression tree, random forest, gradient boosting | Prediction of the solar irradiance |
| [53] | 2016 | Support vector regression (SVR), gradient boosted regression (GBR), Random forest regression (RFR) | Forecasting of solar radiation values |
| [54] | 2010 | Neural networks and fuzzy logic models | Predict half daily values of global solar irradiance |

| [55] | 2018 | Weighted regularized extreme learning machine | Improvement of wind speed prediction accuracy |

## 2.2 ML and DL Applications for Securing the SG

Before moving in to the application side of securing SG using ML and DL, first we discuss about the security objectives and security requirements that every SG must met.

**Smart Grid Security Objectives**

As discussed on [4, 7], following are the three main security objectives that a SmartGrid should meet.

- Confidentiality: This ensures only authorized personnel will have the access to the critical and confidential information which is really necessary for preventing unauthorized disclosure of information.
- Integrity: This ensures data will not be tampered during the storage or the transmission which may lead to incorrect decisions in energy management.
- Availability: This ensures timely and reliable access to information whenever it is necessary and this guarantees the system reliability.

  In order to ensures a timely and reliable delivery of energy these three main key security objectives needed to be met.

**Security Requirements**

Apart from the previously mentioned high level security objectives, following security requirements needed to satisfied which covers most of the aspects of physical, network and the cyber security in the context of SG , which includes physical and massive network infrastructure for energy delivery [4,7].

- Identification and authentication: SG network infrastructure comprises of millions of devices and users, which should be verified and authenticated before granting the access to the network and critical resources.
- Access control: This ensures that resources are only accessed by verified by authorized personnel only.
- Secure communication protocol: As the information that flows contains timely critical information appropriate secure protocols must be devised which is high reliable and secure.

- Attack detection: Compared to the tradition power grid, SG comprises of massive communication network spawns over a large geographic area. Therefore it is impossible to say any part of the network is not vulnerable to network attacks. Therefore mechanisms should be placed to perform filtering, testing and comparison of network traffic to detect and identify any anomalies present.
- Self-healing – In the presence of an attack (e.g.: Denial of Service) network and the other mission critical devices must have the ability of self-healing for continuing its work even under an attack.

These security requirements along with the objectives will empower the smart grid with adequate security capabilities for facing against cyber-attacks and successfully mitigating them.

**ML and DL Applications for Securing Smart Grids**

SG require higher degree of consistent network connectivity to function smoothly. This constant access to network connectivity poses huge potential to open up for new vulnerabilities that comes through the network [6, 9,31,32]. Following Table 1 summarizes the ML and DL applications that is introduced and devised to secure SG.

**Table 2.** Machine Learning Applications for Securing Smart Grid.

| Reference | Year | ML/DL Technique | Application |
| --- | --- | --- | --- |
| [9] | 2019 | Unsupervised machine learning techniques | A scalable anomaly detection engine that is suitable differentiating cyber-attacks |
| [10] | 2014 | Support vector machine | Identifying and detecting stealthy attacks |
| [11] | 2015 | Combination of supervised and semi-supervised techniques | Classification of measurements as being either secure or attacked using the underlying data |
| [12] | 2014 | Artificial neural network | Examine energy consumption data to report energy fraud |
| [13] | 2017 | Wide and deep convolutional neural networks | Analysis on the data for detecting electricity theft |
| [14] | 2018 | Recurrent Neural Networks | Detect False data injection (FDI) attacks on SG's |
| [15] | 2018 | Support vector machine | Detecting covert cyber deception assault from data that are collected through SG communications network |
| [16] | 2017 | Random forest, decision trees, neural networks, support vector machines and naive bayes classification | Intrusion attack detection based on the datasets gathered from synchrophasor devices |
| [17] | 2019 | Isolation forest | Detecting covert data integrity assaults in SG communications networks |
| [18] | 2019 | Recurrent neural networks | Detecting fraudulent transactions in the block chain-based energy network and network attacks |
| [19] | 2017 | Gaussian mixture model | A statistical anomaly detection approach for detecting false data injection attacks |
| [20] | 2015 | Kalman filter estimation ,Chi-square detector and cosine similarity matching | Detection of attacks in smart grids communication systems |
| [21] | 2016 | Artificial neural network | Contextual anomaly detection for SG |
| [22] | 2019 | k-nearest neighbors and support vector machine | Detection of stealthy False data injection attacks |
| [23] | 2014 | Random Forests ,Support vector machine and Naive bayes | Differentiating the types of power system disturbances and detecting cyber-attacks |
| [56] | 2018 | Stacked auto-encoder | Interval State Estimation of AC Smart Grids against Sparse Cyber Attacks |

## 2.3 Other ML and DL Application Ares for SG

The intention of this subsection is to, discuss the rest of the areas that ML and DL can be used other than security and forecasting about energy generations based on the existing literature. In addition in this section, we highlight how this ML and DL used with other enabling technologies like Cloud Computing, Blockchain, Fog computing, Edge Computing, and Big data. Table 3 summarizes the rest of the areas that ML and DL applications are introduced and devised.

**Table 3.** Other ML and DL Application Ares for SG.

| Reference | Year | Application |
|---|---|---|
| [24] | 2020 | A novel Multidirectional Long Short Term Memory (MLSTM) technique to predict the stability of the smart grid network which uses Gated Recurrent Units (GRU), traditional LSTM and Recurrent Neural Networks (RNN) DL techniques |
| [57] | 2019 | A novel deep learning and blockchain-based [26,27] energy framework for SG's which employs recurrent neural networks for detecting network attacks and fraudulent transactions |
| [58] | 2020 | A framework that aims to detect electricity consumption anomalies accurately and timely using sensor processing, smart meter readings, machine learning and block-chain |
| [59] | 2013 | A scalable software platform for the Smart grid CPS using Cloud technologies [34] which contains scalable machine-learning models trained over massive datasets for agile demand forecasting |
| [60] | 2020 | A novel hybrid artificial intelligence classification method focus on the charging behavior profile of Electronic vehicles (EV) based on cloud computing and fog computing. Through this classification method, target EVs can be accurately identified. |
| [61] | 2020 | Provides IoT big data analytics with fog computing for household energy management in smart grids |
| [62] | 2019 | Edge computing data analytics solutions for the smart grid data processing |

## 3    Future directions

ML is undoubtedly a great way to sift through big data and extract meaningful information that can extensively help in the recognition of demand and generation patterns, generation forecasting, control, etc. In current literature, a number of methods have already been presented, and more novel techniques are being worked on for improved accuracy and performance for the time being.

The following are future directions to efficiently deploy big data for proper analysis in the SG. It is noted that comprehensive R&D efforts needed to be done to increase the interoperability between devices, data analytic tools and data repositories.

1. Utilization of heterogeneous data from various sources
   It is believed that in the future there will be advanced applications that can use multiple sources of big data which can help in assessing the dependence of critical infrastructure on power grids. They will effectively help to uncover crucial hidden information and generate meaningful insights.

2. Advancements of predictive algorithms
   In addition to exploiting fine-grained patterns within the data, advanced artificial intelligence based techniques such as DL are important to make the decision process less reliant on human intervention. However underlying Smart grid data will continue to grow with the time. Therefore, the scalability of the ML models will be a huge concern and play a critical role in the future SG data analytics.

3. Integrations with enabling technologies
   For better efficiency, scalability, security, cost reduction, and performance data analysis algorithms will integrate with enabling technologies like Cloud, Blockchain, Fog computing, and Edge computing.

4. Integration with real time control, operation
   Data analytics should be integrated into real time analytics so it provides real time situational awareness in order to run the grid continuously. Therefore future R&D efforts need to put more weight on this.

# 4      Limitations

Analysis of the underlying data is the key to identify meaningful information and resolve many challenging problems that cannot be solved by conventional methods. But because of the intense nature and the massive infrastructure, this ML and DL based applications may have problems like learning from imbalanced data, difficulties in interpretation and difficulties in transfer learning, etc.

When it comes to the security aspect of the Smart grid is still under development and still lot more research carried out since security must be taken in to account with the electrical power systems and it has be incorporated fundamentally when developing smart grid. Smart Grid comprises of reinforced communication network architecture including heterogeneous devices and protocols which ultimately connect with the World Wide Web forming a larger network and it should be highly scalable and secure. The Smart grid therefore needs sound security solution explicitly designed for distinct network applications, making cyber security a very fruitful and demanding research field for the Smart grid in future. Smart grid as a vital infrastructure requires highest degree of protection hence a comprehensive architecture with safety built in from the start is needed.

## 5     Conclusion

Big data analytic in the Smart grid is a new field of research that has drawn greater interest from the government, academia, and industry. In this paper, we surveyed and summarized machine learning and deep learning based applications that are introduced and devised in the context of the Smart grid. The interconnected devices in the Smart grid and the data they generate give rise to the desperate necessities of proper analysis, as the value it holds. Finally, the findings of this study were provided summarizing the key machine learning, deep learning based applications based on the literature and we hope this will assist the researchers in this area.

14